\pdfoutput=1

\documentclass[11pt]{article}

\usepackage{ACL2023}

\usepackage{times}
\usepackage{latexsym}

\usepackage[T1]{fontenc}

\usepackage[utf8]{inputenc}

\usepackage{microtype}

\usepackage{inconsolata}

\usepackage{xcolor,colortbl}
\usepackage{booktabs}
\usepackage{multirow}
\usepackage{graphicx}
\usepackage{amsmath}
\usepackage{pgfplots}
\usepackage{pgfplotstable}
\usepackage{tikz}
\usepackage{enumitem}
\usepackage[normalem]{ulem}
\useunder{\uline}{\ul}{}

\newcommand{\mulam}{\textsc{MuLan}\hspace{0.8pt}\raisebox{-0.3pt}{\includegraphics[width=0.15in]{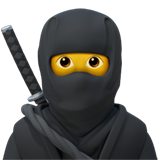}}}
\newcommand{\llamaone}{LLaMA}
\newcommand{\alpaca}{Alpaca}
\newcommand{\llamatwo}{LLama-2}
\newcommand{\llamachat}{LLama-2$_{\text{Chat}}$}
\newcommand{\falcon}{Falcon}
\newcommand{\falconinst}{Falcon$_{\text{Instr}}$}

\definecolor{color_immutable1}{RGB}{106,169,79}
\definecolor{color_immutablen}{RGB}{241,194,50}
\definecolor{color_mutable}{RGB}{166,28,0}

\definecolor{wedge1}{RGB}{ 190  30  46}
\definecolor{wedge2}{RGB}{ 247 148  30}
\definecolor{wedge3}{RGB}{  28 117 188}
\definecolor{wedge4}{RGB}{ 181 212 239}
\definecolor{wedge5}{RGB}{141 199  63}
\definecolor{wedge6}{RGB}{123  82  49}

\title{\mulam{}: A Study of Fact Mutability in Language Models }

\author{Constanza Fierro$^{\dagger}$ \ \ Nicolas Garneau$^{\dagger}$ \ \ \\
        \textbf{Emanuele Bugliarello$^{\natural}$ \ \ Yova Kementchedjhieva$^{\ddagger}$ \ \ Anders S{\o}gaard$^{\dagger}$} \\
        $^{\dagger}$Department of Computer Science, University of Copenhagen \\ 
        $^{\ddagger}$ Mohamed bin Zayed University of Artificial Intelligence \\
        $^{\natural}$Google DeepMind \\ 
        \texttt{\{c.fierro,nicolas.garneau,soegaard\}@di.ku.dk}}

\begin{document}
\maketitle
\begin{abstract}
Facts are subject to contingencies and can be true or false in different circumstances. 
One such contingency is \textit{time}, wherein some facts mutate over a given period, e.g., the president of a country or the winner of a championship.  
Trustworthy language models ideally identify mutable facts as such and process them accordingly. 
We create \mulam{}, a benchmark for evaluating the ability of English language models to anticipate time-contingency, covering both 1:1 and 1:$N$ relations. We hypothesize that mutable facts are encoded differently than immutable ones, hence being easier to update. In a detailed evaluation of six popular large language models, we consistently find differences in the LLMs' confidence, representations, and update behavior, depending on the mutability of a fact. Our findings should inform future work on the injection of and induction of time-contingent knowledge to/from LLMs.\footnote{Code and dataset: \url{https://github.com/coastalcph/fact_mutability}.}

\end{abstract}

\section{Introduction}

Pretrained and large language models (LLMs) trained on vast amounts of text are known to encode factual knowledge~\cite{petroni-etal-2019-language, jiang-etal-2020-x, liu2023pre}. By using cloze-tests or next-token prediction, it is possible to retrieve facts memorized during pretraining \cite{meng-etal-2022-rewire, yin-etal-2022-geomlama, chalkidis-etal-2023-lexfiles}. 
However, factual knowledge changes over time, and facts such as who is the president of a country, or where someone lives, are time-contingent truths, drifting or mutating with the passage of time. 

\citet{dhingra-etal-2022-time} posit that pretrained LLMs are not time-aware, given that they do not predict the right answer for a specific time period.
\citet{jain-etal-2023-language-models} and \citet{qiu2023large} prompted LLMs to measure their ability to reason with temporal information, finding again that LLMs show limited time awareness and lag behind human performance. In this paper, we challenge this belief by studying the representations and behavior of LLMs with respect to facts that mutate over time, hypothesizing that despite the low performance encountered in the temporal tasks studied so far, LLMs do encode mutability in their knowledge representations.

The inherent temporality of facts has also been studied from the perspective of model updates, exploring
how to best edit the knowledge embedded in pretrained LLMs. Most of the editing techniques have focused on modifying the parameters directly \citep{de-cao-etal-2021-editing,mitchell2022fast, meng2022locating}, however \citet{cohen2023evaluating} found that a simple in-context editing obtains more consistent updates than direct parameter modifications. In this paper, we explore LLMs behavior with respect to mutability by studying the effectiveness of knowledge updates given a fact's mutability type.

In order to  study LLMs' ability to anticipate when facts are time-contingent, we introduce the \mulam{} (\textbf{Mu}tability in \textbf{Lan}guage models) dataset. Unlike LAMA \cite{petroni-etal-2019-language}, which contains mostly immutable relations, and TempLAMA \cite{dhingra-etal-2022-time}, which contains exclusively mutable relations, \mulam{} contains a balanced mix of both types of relations, curated to enable the controlled study of mutability, while also disentangling this phenomenon from cardinality, i.e., the property of some relations to take multiple objects at the same time. \mulam{} contains 35 relations extracted from WikiData,\footnote{\url{https://www.wikidata.org/}.} each with up to 1,500 queries.
Equipped with this resource, we address the following research questions.

{\bf RQ1:} Will LLMs exhibit lower confidence and performance on time-contingent truths? We find that there is a difference in performance, but the gap in confidence is even more impressive (Table~\ref{tab:f1_scores_summary}). 

{\bf RQ2:} Will LLMs represent time-contingent truths differently, making it easy to differentiate representations in terms of mutability? Using probe classifiers, we find that representations do encode mutability (as shown in Table~\ref{tab:mdl}).

{\bf RQ3:} Will updating mutable facts be easier than immutable ones? Using in-context learning, we show that mutable facts are indeed more consistently updated than immutable ones (see Table~\ref{tab:updates}). 

In the study of RQ1-3, we probe six popular LLMs on \mulam{}. Analyzing their predictions, representations, and update behavior, we find consistent differences depending on a fact's mutability. 
This indicates that while time awareness in LLMs cannot be detected through prompting \cite{dhingra-etal-2022-time,jain-etal-2023-language-models,qiu2023large}, it is present in their representations.
This finding should inform the design of methods for the induction of factual knowledge from LLMs, and for updating knowledge contingent with the passage of time. 

\section{The \mulam~Benchmark}

\paragraph{Probing} Similar to \citet{petroni-etal-2019-language}, we probe language models for their encoded knowledge using Wikidata triples (\texttt{subject}, \texttt{relation}, \texttt{object}) e.g., (\texttt{Germany}, \texttt{capital}, \texttt{Berlin}).
To probe language models, these triples can be formulated as a query for which, given a subject and a relation (a query; e.g., ``The capital of \textit{Germany} is \textit{[X]}.''), the goal is to retrieve the corresponding object(s).
Naturally, some relations, e.g., {\tt capital of}, are binary or one-to-one (1:1)\footnote{Each subject has only one correct object.} but many others may have many possible completions being $n$-ary (1:N), e.g., the relation \texttt{shares borders with}. Both 1:1 and 1:N relations can be involved in time-contingent truths.\footnote{For example, \texttt{Gianluigi Buffon} was playing for \texttt{Paris Saint-Germain} and \texttt{Italy's national association football team} in 2018, but {\em not} in 2023.} But time-contingent 1:1 facts are, in a sense, 1:N across time, so including 1:N facts that are {\em not} time-contingent provides for an interesting control scenario. Therefore, we construct \mulam{} to contain 3 sets of relations: Immutable-1 (1:1), Immutable-N (1:N), and Mutable (1:N). Note how instances of Immutable-N and Mutable relations might look similar to a language model from a training perspective, given that the data is shuffled at training time.\footnote{For example, the data may contain a sentence about France neighboring Germany, and another about neighboring the Netherlands. It may also contain a sentence about Buffon playing for PSG, and another playing for Juventus.}

\paragraph{Dataset} To create \mulam{}, we first select the set of relations for each mutability type. We use LAMA~\citep{petroni-etal-2019-language} and TempLAMA~\citep{dhingra-etal-2022-time} as a starting point and further expand the number of candidate relations with the WikiData database.
We then verify the mutability and cardinality of a relation by using the average number of objects, defining a threshold to differentiate between Immutable-1 and Immutable-N, and validating that the cardinality of Mutable and Immutable-N is similar on average. 
By doing so, we obtain 12 Immutable-1 relations, 10 Immutable-N relations, and 13 Mutable relations.
All relations are listed in Table~\ref{tab:relations} in Appendix~\ref{app:dataset}.
For each relation, we retrieve up to 1,500 widely known subjects.\footnote{We use the number of translated pages on Wikipedia as a proxy for widely known entities.} For each query, we retrieve the corresponding objects along with their aliases (e.g., New York--NYC). The dataset consists of 47k subject--relation queries (17k Immutable-1, 13k Immutable-N, and 17k Mutable).
For each relation, we create 5 paraphrased templates to account for prompt brittleness \citep{elazar-etal-2021-measuring}.
We alter ambiguous templates to specify the desired entity type, e.g., we change the LAMA template for relation P19 (place of birth) from ``[X] was born in [Y]'' to ``[X] was born in the location of [Y].'', to remove the year vs. location ambiguity.

\paragraph{Models} We consider two types of pretrained LLMs: foundation and instruction-tuned models.
As foundation models, we use \llamaone{}~\cite{Touvron2023LLaMAOA}, \llamatwo{} \citep{touvron2023llama2}, and \falcon{}~\citep{falcon}; we evaluate the 7 billion parameters versions for all the models.
\llamaone{} is a model designed for general-purpose tasks such as question answering, and text generation. The \llamatwo{} model is the more recent and more performant version, which is better suited for reasoning tasks.
Finally, \falcon{} has been trained on RefineWeb, an enhanced curated corpora~\citep{refinedweb}.
We also consider their instruction-tuned variants, namely \alpaca{}~\cite{alpaca}, \llamachat{}~\citep{touvron2023llama2}, and \falconinst{}.\footnote{For the instruction-tuned models, we add the corresponding format (used during fine tuning) to the query. We use the same instructions for all of them. For the foundation models we only use the query without any specific formatting or instruction.}
We hypothesize that instruction-tuned models will be inclined to produce more succinct answers, making them easier to evaluate in an open-generation setup such as the one we consider.

\paragraph{Evaluation}  
We perform greedy decoding with a beam of size 1 for all models, and we follow the evaluation scheme of ~\citet{rajpurkar-etal-2016-squad} measuring the average word overlap between the prediction and ground truth answer (the list of WikiData objects along with their aliases).
In case of multiple answers, we take the maximum F1 score.\footnote{We normalize both the prediction and ground truths to remove punctuation, articles, upper cases, and extra white spaces. We also truncate the prediction (after normalization) to have at most the length of the target, to account for foundation models' longer predictions.}

\section{Results}\label{sec:results}

\paragraph{Performance \& Confidence (RQ1)}\label{mulam_results} Table~\ref{tab:f1_scores_summary} shows how LLMs are, in general, better at predicting immutable facts than mutable ones.
\llamatwo{} and \llamachat{} show strong performance across relation types.
Relations \texttt{native language} (P130) and \texttt{located in} (P30) are the easiest to predict, presumably because their output space is rather limited (detailed results in App.~\ref{app:f1_scores}).
Next, we consider models' confidence across mutable and immutable facts, where we define the confidence of a model as being the probability of the first predicted token. Table~\ref{tab:f1_scores_summary} shows, across models, a significant difference in confidence between predicting Immutable-1 facts and predicting Mutable facts. The drop from Immutable-1 to Immutable-N facts is unsurprising given that there are several plausible answers \citep{holtzman-etal-2021-surface}, but the additional drop from Immutable-N to Mutable suggests that LLMs might treat mutable facts differently.
Figures in Appendix~\ref{sec:appendix} show how slightly correlated the F1 score is with the confidence across models.

\begin{table}[t!]
    \centering\scriptsize\setlength{\tabcolsep}{3pt}
    \resizebox{0.48\textwidth}{!}{
    \begin{tabular}{lcccccc}
    \toprule
    & \multicolumn{2}{c}{\textbf{Immutable-1}} & \multicolumn{2}{c}{\textbf{Immutable-N}} & \multicolumn{2}{c}{\textbf{Mutable}}\\ 
    \cmidrule(lr){2-3}
    \cmidrule(lr){4-5}
    \cmidrule(lr){6-7}
    \textbf{Models} & \textbf{F1} & \textbf{Conf.} & \textbf{F1} & \textbf{Conf.} & \textbf{F1} & \textbf{Conf.}\\
    \midrule
    \llamaone{} & 55.1 & \cellcolor[gray]{0.9}0.41 & 48.5 & \cellcolor[gray]{0.9}0.36 & 24.6 & \cellcolor[gray]{0.9}0.27\\
    \alpaca{} & 53.1 & \cellcolor[gray]{0.9}0.75 & 49.9 & \cellcolor[gray]{0.9}0.66 & 31.8 & \cellcolor[gray]{0.9}0.54\\
    \llamatwo{} & \textbf{59.1} & \cellcolor[gray]{0.9}0.55 & 50.7 & \cellcolor[gray]{0.9}0.41 & 26.6 & \cellcolor[gray]{0.9}0.28\\
    \llamachat{} & 56.4 & \cellcolor[gray]{0.9}0.88 & \textbf{50.9} & \cellcolor[gray]{0.9}0.83 & \textbf{32.1} & \cellcolor[gray]{0.9}0.79\\
    \falcon{} & 50.9 & \cellcolor[gray]{0.9}0.37 & 45.2 & \cellcolor[gray]{0.9}0.26 & 17.3 & \cellcolor[gray]{0.9}0.19\\
    \falconinst{} & 48.2 & \cellcolor[gray]{0.9}0.48 & 45.2 & \cellcolor[gray]{0.9}0.39 & 24.0 & \cellcolor[gray]{0.9}0.28\\
    \bottomrule
    \end{tabular}
    }
    \caption{Average F1 score and confidence of each model across relation types in \mulam.}
    \label{tab:f1_scores_summary}
\end{table}





\tikzstyle{immutable1}=[thick, solid, #1, mark size=1.2pt, mark=*, mark options={#1, solid}]
\tikzstyle{immutablen}=[thick, #1, dashed, mark size=1.2pt, mark=*, mark options={#1, solid}]
\tikzstyle{mutable}=[thick, #1, dotted, mark size=1.2pt, mark=*, mark options={#1, solid}]
\def\height{5.5cm}

\paragraph{Representations Encode Mutability (RQ2)}\label{probe_clf}
We leverage \mulam{} to study how separable the mutable class is. Hence, we train two probe classifiers\footnote{We study Mutable vs. Immutable-1, and Mutable vs. Immutable-N separately.}
~\citep{hewitt-liang-2019-designing}
and define a control task where the labels (mutable or immutable) are randomly assigned to each relation. Good performance under such perturbation would indicate memorization of spurious correlations. 
The quality of a probe classifier is commonly measured w.r.t. accuracy, however, this does not account for how difficult (more complex probe) it is to attain such performance. Therefore, we use Minimum Description Length (MDL; \citealt{voita-titov-2020-information}), which measures the minimum number of bits needed to transmit the correct labels \(\{y_i\}_{i=1...n}\) for each example \(\{x_i\}_{i=1...n}\). Specifically, we compute the online codelength \citep{rissanen-online-cl}, where both parts agree before the transmission on: the model family \(p_\theta (y|x)\), a set of learnable parameters \(\theta\), some initial random seeds, the optimization algorithm, and the timesteps \(t_0 < t_1 < ... < t_S = 100\%\) to send the data in batches. The first block of data (\(x_{[1:t_0]})\) is transmitted with a uniform code,\footnote{The uniform codelength sends all the labels without using any probabilistic model, so the probability of each label is uniform \(p(y|x) = 1/K\) yielding \(L_{unif}(y_{1:n}|x_{1:n})=nlog_2K\).} and from then on at time step \(i\) both parts train a model on the already transmitted data obtaining the same model \(p_{\theta_i}\), and the next block of labels \(x_{]t_i:t_{i+1}]}\) is transmitted using the predictions of \(p_{\theta_i}\). The online description length for \(K\) possible labels is:
\begin{align}
    &L_\text{online}(y_{1:n|x_{1:n}}) = t_0\log_2 K \nonumber \\
    &- \sum_{i=1}^{S-1} \sum_{j=t_i+1}^{t_{i+1}} \log_2 p_{\theta_i}(y_j | x_j).
\end{align}
\noindent
Then, the compression is defined as how much the online codelength has been compressed in comparison to transmitting all the labels with a uniform codelength, that is, \(L_\text{uniform} / L_\text{online}\).

To train the probe classifier, we select disjoint sets of relations to be used as train, validation, and test; such that we can measure generalization to other types of subjects and objects.\footnote{We remove queries with overlapping subjects across training splits. The full list of relations is in Appendix \ref{appendix:probe}.} We encode the triplet using one of the 5 available templates and use the last token representation from the last layer. For the training and validation data (used for computing the MDL) we preprocess \mulam{} so as to only use one template per triplet and only one object (both chosen randomly). For the test data (all the remaining relations) we use all the 5 templates and only one object (chosen randomly); we report the macro average across relations. Note that for the control task it does not make sense to look at the probe accuracy in the validation or test set, as these have a different set of relations than the training (and so there is no connection between the random labels assigned for train and test).

We experiment with a linear projection as the probe model, which uses as input the last token representation from the last layer (details in \S\ref{appendix:mdl}). 
Table \ref{tab:mdl} presents the results for the classification task of Mutable vs Immutable-1 and Mutable vs Immutable-N. We see that the codelength of the mutable class is much smaller than that of the random labels, and the compression is above that of a uniform codelength. This means that it is relatively {\em easy} to tell mutable facts from immutable facts, supporting our findings in RQ1. Accuracy is generally high, which validates that the probe can find dimensions that encode mutability across unseen types of relations. To further measure whether these probes are relying only on spurious features of the templates themselves, we train a Naive Bayes classifier on bag of words of the templates, obtaining an accuracy of 0.64 for Immutable-1 and 0.6 for Immutable-N. Given that the accuracy of all probes is higher, we conclude that the probes found mutability features in the representations. Note that higher accuracy could be obtained by experimenting with other probe architectures and performing a more exhaustive search of hyperparameters; however, our main metric is the MDL, and we already find with this result that mutability is more salient than simply template---or type---representations.

Additionally, we assess whether these mutability features found in the representations might simply be frequency features. We study the accuracy of the classifier by dividing the test set in frequency bins,\footnote{We use the number of Wikipedia pages of an entity as a proxy of frequency for its facts.} the plots can be found in \S\ref{appendix:freq_prob_plots}. The main observation is that the performance of the classifier cannot be explained simply by frequency, e.g. in the 3rd bin there are 964 Immutable-1 and 840 Mutable examples, and the Immutable-1 classifiers (Figure \ref{fig:acc_bins_classifier_1_1}) get around 85-95\% accuracy (depending on the model), way above a classifier using simply frequency (where low frequent examples would be considered mutable, therefore getting ~46\% accuracy); the same goes for higher bins and for the Immutable-N classifiers (see \S\ref{appendix:freq_prob_plots}).




\begin{table}[t]
\centering\scriptsize\setlength{\tabcolsep}{3pt}
\resizebox{0.48\textwidth}{!}{
\begin{tabular}{@{}lllll@{}}
\toprule
\textbf{Data} & \textbf{Model} & \textbf{Acc.} & \multicolumn{1}{c}{\textbf{Codelength}} & \multicolumn{1}{c}{\textbf{Compr.}} \\ \midrule
\multirow{6}{*}{Imm-1} & \llamaone{} & 0.86 & 610 / 4880 & 10.2 / 1.2 \\
 & \alpaca{} & 0.87 & 487 / 6627 & 12.7 / 0.9 \\
 & \llamatwo{} & 0.85 & 537 / 4722 & 11.5 / 1.3 \\
 & \llamachat{} & 0.85 & 778 / 4620 & 8.0 / 1.3 \\
 & \falcon{} & 0.89 & 649 / 4694 & 9.6 / 1.3 \\
 & \falconinst{} & 0.89 & 617 / 4329 & 10 / 1.4 \\ \midrule
\multicolumn{1}{r}{\multirow{6}{*}{Imm-N}} & \llamaone{} & \multicolumn{1}{r}{0.78} & \multicolumn{1}{r}{1058 / 3354} & \multicolumn{1}{r}{6.4 / 2.0} \\
\multicolumn{1}{r}{} & \alpaca{} & 0.88 & 453 / 901 & 15.0 / 7.5 \\
\multicolumn{1}{r}{} & \llamatwo{} & 0.87 & 1014 / 2513 & 6.7 / 2.7 \\
\multicolumn{1}{r}{} & \llamachat{} & 0.74 & 764 / 4671 & 8.9 / 1.4 \\
\multicolumn{1}{r}{} & \falcon{} & 0.83 & 714 / 1999 & 9.5 / 3.4 \\
\multicolumn{1}{r}{} & \falconinst{} & 0.87 & 534 / 3302 & 12.7 / 2.0 \\ \bottomrule
\end{tabular}
}
\caption{Representations classification results between immutable and mutable. The MDL results are presented as: mutability / random relation labels. Accuracy is the macro average on the test relations.}
\label{tab:mdl}
\end{table}



\begin{table}[t]
\centering\scriptsize\setlength{\tabcolsep}{3pt}
\resizebox{0.48\textwidth}{!}{
\begin{tabular}{@{}lccc@{}}
\toprule
\textbf{} & \multicolumn{3}{c}{\textbf{Update Success Rate}} \\ \cmidrule(l){2-4} 
\textbf{Model} & \textbf{Immutable-1} & \textbf{Immutable-N} & \textbf{Mutable} \\ \midrule
\llamaone{} & \textbf{0.718} & 0.650 & 0.588 \\
\alpaca{} & 0.734 & \textbf{0.862} & {\ul 0.845} \\
\llamatwo{} & 0.566 & 0.654 & \textbf{0.725} \\
\llamachat{} & 0.397 & 0.393 & \textbf{0.526} \\
\falcon{} & 0.449 & 0.382 & \textbf{0.610} \\
\falconinst{} & 0.302 & 0.331 & \textbf{0.528} \\ \bottomrule
\end{tabular}
}
\caption{Fraction of successful knowledge updates.}
\label{tab:updates}
\end{table}

\paragraph{Edits (RQ3)} 
Finally, we study how mutability affects knowledge updates in LLMs.
To do so, we first select a set of queries per model, each set is composed of facts that an LLM has memorized.\footnote{Facts answered perfectly and with high confidence.} From this set, we then sample the queries so as to have equal number of examples for each mutability type.\footnote{We sample uniformly per relation, maintaining as even as possible the number of examples across relations. The final number of examples per model is specified in Appendix \ref{appendix:updates}.}
For each query, we randomly select a new target object to replace the memorized one.\footnote{We sample from the set of correct predictions in the same relation, to account for definite articles or prepositions.} To update the knowledge of the LLMs we follow \citet{cohen2023evaluating} and \citet{sogaard-2021-lockes}, and we convert the queries to be of the form ``\textit{Imagine that {\tt <fact\_update>}. Then, {\tt <query>}}.'' 
Here, {\tt <query>} uses the best performing template, and {\tt <fact\_update>} contains one of the other 4 templates with the new object. We consider an update to be successful if the model's top prediction is an exact match to the new target object. 
Table \ref{tab:updates} shows that, for most of the models, the rate of Mutable facts successfully updated is consistently {\em higher} than for Immutable ones.
We find a slightly different trend for \llamaone{}, where the pretrained version is more effective at updating Immutable-1 and the instruct-tuned version obtains similar numbers for Immutable-N and Mutable.

We also analyze whether this different behavior might be due only to differences in frequency. We break down the updates accuracy in the three highest frequency percentiles. As we can see in Table~\ref{tab:acc_scores_freq}, there is a clear trend across models that mutable facts, even if they are frequent, are easier to update by models, whereas immutable ones (either 1 or N) are harder. In lower bins there is no clear trend in accuracy differences between mutability types.

\begin{table}[t!]
    \centering\scriptsize\setlength{\tabcolsep}{2pt}
    \resizebox{0.48\textwidth}{!}{
    \begin{tabular}{lccccccccc}
    \toprule
    & \multicolumn{3}{c}{\textbf{Immutable-1}} & \multicolumn{3}{c}{\textbf{Immutable-N}} & \multicolumn{3}{c}{\textbf{Mutable}}\\ 
    \cmidrule(lr){2-4}
    \cmidrule(lr){5-7}
    \cmidrule(lr){8-10}
    \textbf{Models} & \textbf{1st} & \textbf{2nd} & \textbf{3rd} & \textbf{1st} & \textbf{2nd} & \textbf{3rd} & \textbf{1st} & \textbf{2nd} & \textbf{3rd} \\
    \midrule
    \llamaone{} & 71.4 & \underline{80.9} & \textbf{82.3} & 65.0 & {78.9} & 30.7 & {51.7} & {52.1} & 46.6 \\
    \alpaca{} & 76.8 & {82.1} & 77.8  & 82.4 & 87.1 & \textbf{95.3} & \underline{91.1} & 85.4 & 81.4  \\
    \llamatwo{} & {62.6} & 56.2 & 36.3 & 66.1 & 51.7 & \textbf{82.3} & 70.0 & \underline{72.1} & 62.1 \\
    \llamachat{} & {41.1} & 41.0 & 40.2 & 38.9 & 39.6 & {42.2} & 42.6 & \textit51.2 & \textbf{51.5}\\
    \falcon{} & 42.0 & \underline{63.2} & 42.8 & 30.6 & 46.2 & {57.1} & 56.7 & \textbf{69.7} & 60.0\\
    \falconinst{} & 31.8 & {37.1} & 23.2 & 27.6 & 42.3 & {44.0} & 46.9 & \underline{55.5} & \textbf{56.7} \\
    \bottomrule
    \end{tabular}
    }
    \caption{Update accuracy according to the 1st, 2nd, and 3rd highest percentile in terms of entities' frequency. In bold/underline the highest/2nd highest accuracy per model. The results show that even when comparing frequent entities, mutable facts are easier to update.}
    \label{tab:acc_scores_freq}
\end{table}

\section{Conclusion}
Using our novel resource, \mulam{}, we find new evidence for time awareness in LLMs, specifically for encoding of fact mutability in their representations and for the comparative ease of editing of mutable facts versus immutable ones. Research on the learning dynamics of LLMs may further investigate how LLMs trained on shuffled data acquire such time awareness. From a practical standpoint, on the other hand, our findings should inform future work on the induction of knowledge from LLMs, and the design choices for updating LLMs.


\section*{Limitations}

The accuracy of \mulam{} is dependent on the accuracy of Wikidata, since we extracted the queries using its API.
Also, \mulam{} and the models analyzed are exclusively in English.
Even though we try to control and analyze confounds, there might be other possible confounds, e.g. objects' types in the updates, and a more exact estimation of entities' frequency in the training data.
The experiment we perform regarding the updates of mutable facts is limited to contextual modification of LLM knowledge, and we hope that this preliminary experiment will pave the way for other update mechanisms.



\bibliography{anthology,custom}
\bibliographystyle{acl_natbib}

\clearpage
\appendix

\section{The \mulam{} Dataset}\label{app:dataset}

Table~\ref{tab:relations} presents all the relations included in the \mulam{} dataset, and some concrete examples can be found in Table~\ref{tab:queries_examples}.

We also provide the performance and confidence results break down per relation in Table~\ref{tab:f1_scores}. We note that some of the mutable examples in \mulam{} only have one target answer, despite having multiple potential answers in reality. This limitation arises from the incomplete nature of WikiData. We compute the average F1-scores when discarding these incomplete examples and the models obtain: 30.19 \llamaone{}, 40.20 \alpaca{}, 32.6 \llamatwo{}, 40.80 \llamachat{}, 21.91 \falcon{}, 31.60 \falconinst{}. We note that the same trends discussed in \S\ref{sec:results} still stand.

\subsection{Frequency Per Mutability Type}\label{appendix:mulan_freq}
We use the number of Wikipedia sites (in different languages) of a subject as a proxy of frequency (as done for selecting the subjects, see footnote 6), given that we don’t have access to each model’s training data.\footnote{Note that computing co-occurrences may lead to incorrect conclusions since WikiData is not thorough with all the valid completions over time, so subject-object co-occurrences of mutable facts would be less than in reality because we wouldn't be using all the valid object completions.} Given this frequency proxy, we find that in \mulam{} immutable facts are more frequent than mutable, with average and standard deviation counts of: 
\begin{itemize}
    \item Immutable-1: 87.1 \(\pm\) 64.7
    \item Immutable-N: 103.2 \(\pm\) 64.8
    \item Mutable: 65.4 \(\pm\) 53.7
\end{itemize}

\begin{table*}[t]
\centering\scriptsize\setlength{\tabcolsep}{3pt}
\resizebox{0.98\textwidth}{!}{
\begin{tabular}{r l r |r l r| r l r}
\toprule
\cellcolor{color_immutable1!20}\textbf{PID} & \multicolumn{1}{c}{\cellcolor{color_immutable1!20}\textbf{Immutable-1 Relation}} & \multicolumn{1}{c}{\cellcolor{color_immutable1!20}\textbf{\# O}} & \cellcolor{color_immutablen!20}\textbf{PID} & \multicolumn{1}{c}{\cellcolor{color_immutablen!20}\textbf{Immutable-N Relation}} & \multicolumn{1}{c}{\cellcolor{color_immutablen!20}\textbf{\# O}} & \cellcolor{color_mutable!20}\textbf{PID} & \multicolumn{1}{c}{\cellcolor{color_mutable!20}\textbf{Mutable Relation}} & \multicolumn{1}{c}{\cellcolor{color_mutable!20}\textbf{\# O}} \\
\cmidrule(lr){1-3} \cmidrule(lr){4-6} \cmidrule(lr){7-9}
P19 & place of birth & 1,06 (0.31) & P27 & country of citizenship & 1,35 (0.64) & P6 & head of government & 3,63 (6.86) \\
P20 & place of death & 1,08 (0.30) & P47 & shares border with & 6,39 (4.72) & P39 & position held & 4,18 (5.02) \\
P30 & continent & 1,08 (0.42) & P69 & educated at & 2,39 (1.30) & P54 & member of sports team & 7,17 (4.61) \\
P36 & capital & 1,11 (0.61) & P101 & field of work & 2,37 (1.76) & P108 & employer & 2,15 (1.67) \\
P103 & native language & 1,06 (0.36) & P136 & genre & 2,98 (2.49) & P210 & party chief representative & 2,34 (2.64) \\
P138 & named after & 1,28 (0.80) & P166 & award received & 7,56 (10.90) & P264 & record label & 2,60 (2.49) \\
P140 & religion or worldview & 1,27 (0.79) & P190 & twinned administrative body & 9,13 (8.20) & P286 & head coach & 3,59 (5.88) \\
P159 & headquarters location & 1,20 (0.76) & P530 & diplomatic relation & 25,85 (35.44) & P451 & unmarried partner & 1,95 (3.45) \\
P364 & original language (e.g., film) & 1,24 (0.72) & P1303 & instrument & 1,79 (1.22) & P488 & chairperson & 2,85 (5.24) \\
P449 & original broadcaster & 1,24 (1.07) & P1412 & languages spoken & 1,59 (0.89) & P551 & residence & 1,97 (1.99) \\
P495 & country of origin & 1,19 (0.76) &  &  &  & P937 & work location & 1,94 (2.09) \\
P740 & location of formation & 1,02 (0.14) &  &  &  & P1037 & director / manager & 2,20 (3.50) \\
 &  &  &  &  &  & P1308 & officeholder & 2,91 (7.06) \\
 \bottomrule

\end{tabular}
}
\caption{Relations considered in \mulam{} for the Immutable-1, Immutable-N, and Mutable types along with their Wikidata identifier (PID) and average number of objects per query (\#O) with their respective standard deviation.}
    \label{tab:relations}

\end{table*}

\begin{table*}[]
    \centering
    \begin{tabular}{>{\centering\arraybackslash}p{4cm}|>{\centering\arraybackslash}p{6.5cm}|>{\centering\arraybackslash}p{4cm}}
    \toprule
        \textbf{Relation} & \textbf{Query} & \textbf{Objects} \\
    \midrule
        P19 -- Place of birth & \textbf{Aristotle} is originally from \_X\_. & [Stageira]\\
        \cmidrule(lr){1-3}
        P47 -- Shares border with & \textbf{Chile} shares border with \_X\_. & [Argentina, Bolivia, Peru] \\
        \cmidrule(lr){1-3}
        P286 -- Head coach & The head coach of \textbf{Rafael Nadal} is \_X\_. & [Toni Nadal, Carlos Moyá, Francisco Roig] \\
    \bottomrule
    \end{tabular}
    \caption{Examples of queries with their respective list of possible objects.}
    \label{tab:queries_examples}
\end{table*}

\section{Performance of Each Model}
\label{app:f1_scores}

Table~\ref{tab:f1_scores} presents the performance results of each model per relation. To run each model we use one NVIDIA A100-40GB, and it takes from a few hours to 1 day depending on the model.

\begin{table}[t]
    \centering\scriptsize\setlength{\tabcolsep}{4pt}
    \resizebox{0.48\textwidth}{!}{
\begin{tabular}{crcccccc}
\toprule
 &  & \multicolumn{6}{c}{\textbf{Models}}\\
 \cmidrule(lr){3-8}
 \multicolumn{2}{r}{\textbf{Relation}} & \multicolumn{2}{c}{\textbf{Llama}} & \multicolumn{2}{c}{\textbf{LLaMA-2}} & \multicolumn{2}{c}{\textbf{Falcon}}\\
\cmidrule(lr){2-8}
\multirow{12}{*}{\rotatebox{90}{Immutable}} & P19 & 39.57 & 31.01 & \textbf{53.94} & 38.41 & 37.28 & 20.91\\
 & P20 & 15.56 & 36.93 & 29.67 & \textbf{45.02} & 7.64 & 15.75\\
 & P30 & 92.31 & 92.31 & \textbf{92.73} & 88.39 & 88.91 & 85.60\\
 & P36 & 61.57 & 57.35 & \textbf{76.43} & 63.15 & 58.99 & 59.41\\
 & P103 & \textbf{87.25} & 85.73 & 86.55 & 78.55 & 83.68 & 84.11\\
 & P138 & 34.82 & 27.37 & 35.50 & \textbf{44.51} & 26.81 & 16.71\\
 & P140 & 41.55 & 42.50 & 41.88 & 43.30 & 39.85 & \textbf{44.04}\\
 & P159 & 52.21 & 58.69 & 62.18 & \textbf{63.66} & 55.13 & 53.72\\
 & P364 & 84.56 & 76.98 & 86.10 & 58.15 & \textbf{88.82} & 80.85\\
 & P449 & \textbf{52.34} & 41.61 & 52.27 & 49.28 & 36.46 & 34.25\\
 & P495 & \textbf{64.50} & 44.33 & 55.27 & 53.69 & 52.40 & 51.75\\
 & P740 & 34.54 & 42.08 & 36.09 & \textbf{50.53} & 34.49 & 31.39\\
  \cmidrule(lr){2-8}
 \multicolumn{2}{r}{\textbf{Average}} & 55.07 & 53.08 & \textbf{59.05} & 56.39 & 50.87 & 48.21\\
 \cmidrule(lr){1-8}
 \multirow{10}{*}{\rotatebox{90}{Immutable-N}} & P27 & 79.85 & 77.69 & 79.58 & \textbf{80.05} & 78.02 & 76.35\\
 & P47 & 30.20 & \textbf{32.40} & 30.49 & 30.32 & 27.60 & 24.34\\
 & P69 & 52.47 & \textbf{54.04} & 53.38 & 48.19 & 47.20 & 43.15\\
 & P101 & 30.96 & 43.90 & 37.14 & \textbf{55.49} & 22.50 & 31.59\\
 & P136 & 43.63 & 47.61 & 44.79 & \textbf{48.55} & 45.76 & 39.78\\
 & P166 & 35.07 & \textbf{37.34} & 34.74 & 33.81 & 29.50 & 30.14\\
 & P190 & \textbf{28.05} & 16.12 & 27.84 & 8.69 & 27.58 & 26.34\\
 & P530 & 66.39 & 60.96 & 70.83 & \textbf{73.24} & 61.76 & 64.92\\
 & P1303 & 38.54 & 40.61 & 37.85 & \textbf{41.39} & 39.37 & 30.00\\
 & P1412 & 79.37 & 88.39 & \textbf{89.80} & 89.11 & 72.75 & 85.56\\
  \cmidrule(lr){2-8}
 \multicolumn{2}{r}{\textbf{Average}} & 48.45 & 49.90 & 50.65 & \textbf{50.88} & 45.21 & 45.22\\
 \cmidrule(lr){1-8}
 \multirow{13}{*}{\rotatebox{90}{Mutable}} & P6 & 1.73 & 4.25 & 2.56 & \textbf{10.51} & 1.73 & 5.30\\
 & P39 & 55.56 & 55.34 & 58.48 & \textbf{61.05} & 44.56 & 46.53\\
 & P54 & 32.59 & 48.52 & \textbf{53.62} & 42.53 & 31.85 & 45.14\\
 & P108 & 38.45 & \textbf{45.25} & 43.38 & 42.91 & 40.60 & 38.29\\
 & P210 & 7.01 & \textbf{13.95} & 7.18 & 11.69 & 1.81 & 8.94\\
 & P264 & 45.75 & \textbf{45.90} & 42.28 & 42.11 & 15.90 & 30.38\\
 & P286 & 14.88 & \textbf{24.60} & 23.11 & 24.10 & 11.13 & 17.14\\
 & P451 & \textbf{24.08} & 21.97 & 21.04 & 23.59 & 12.98 & 14.50\\
 & P488 & 7.71 & \textbf{22.57} & 6.85 & 21.25 & 3.88 & 13.08\\
 & P551 & 26.52 & 35.73 & 23.87 & \textbf{36.03} & 24.48 & 31.13\\
 & P937 & 47.10 & 49.17 & 30.01 & \textbf{50.51} & 16.72 & 38.72\\
 & P1037 & 12.94 & \textbf{18.68} & 18.51 & 17.25 & 14.16 & 11.43\\
 & P1308 & 5.78 & 27.46 & 14.96 & \textbf{33.89} & 4.43 & 11.33\\
   \cmidrule(lr){2-8}
 \multicolumn{2}{r}{\textbf{Average}} & 24.62 & 31.80 & 26.60 & \textbf{32.11} & 17.25 & 23.99\\
 \cmidrule(lr){1-8}
 \multicolumn{2}{r}{\textbf{mAverage}} & 41.94 & 44.32 & 44.66 & \textbf{45.85} & 36.84 & 38.42\\
\bottomrule
\end{tabular}
}
    \caption{F1-scores of each foundation model (first column) and their instruction counterpart (second) on every relation in \mulam, broken down by query type i.e., Immutable-1, Immutable-N and Mutable and the overall Macro Average.}
    \label{tab:f1_scores}
\end{table}

\section{F1-Score vs Confidence for each model}
\label{sec:appendix}

Correlation analysis between the F1 score and the confidence of each model.
There is a slight trend in the performance of models with respect to their confidence, but the difference in confidence between 0 and 1 is not huge.

\begin{figure}[t]
\centering
\begin{tikzpicture}
\def\espacegraphique{45pt}
\begin{axis}[
    width=\linewidth, height=\height,
    name=int_cosine,
    grid=major, grid style={dashed,gray!50},
    xlabel=F1-Score, ylabel=Confidence,
    ymin=0.0, ymax=1.0, xmin=0.0, xmax=1.1,
    ytick={0.0,0.2,...,1.0},
    xtick={0.0,0.2,...,1.0},
    axis y line*=left, axis x line*=bottom,
    y tick label style={
        /pgf/number format/fixed,
        /pgf/number format/fixed zerofill,
        /pgf/number format/precision=2
    },
    legend pos=south east,
    legend cell align=left,
    legend columns=3,
    transpose legend,
    legend style={draw=none,outer sep=0pt,inner sep=0pt,
    fill=none, font=\footnotesize, yshift=0, xshift=5pt,
        /tikz/column 2/.style={column sep=5pt}
        },
    legend entries={
        Immutable-1,
        Immutable-N,
        Mutable
        }
    ]
    \addlegendimage{no markers, color_immutable1, solid, thick}
    \addlegendimage{no markers, color_immutablen, dashed, thick}
    \addlegendimage{no markers, color_mutable, dotted, thick}
    \addplot+[immutable1=color_immutable1] table[x=f1, y=immutable, col sep=comma]
    {data/f1_confidence_llama_chat.csv};
    \addplot+[immutablen=color_immutablen] table[x=f1, y=immutable_n, col sep=comma]
    {data/f1_confidence_llama_chat.csv};
    \addplot+[mutable=color_mutable] table[x=f1, y=mutable, col sep=comma]
    {data/f1_confidence_llama_chat.csv};

\end{axis}

\end{tikzpicture}
\caption{Average confidence per F1-Score for Llama Chat over all the queries in \mulam. 
}
\label{fig:f1_confidence_llama_chat}
\vspace{-3mm}
\end{figure}
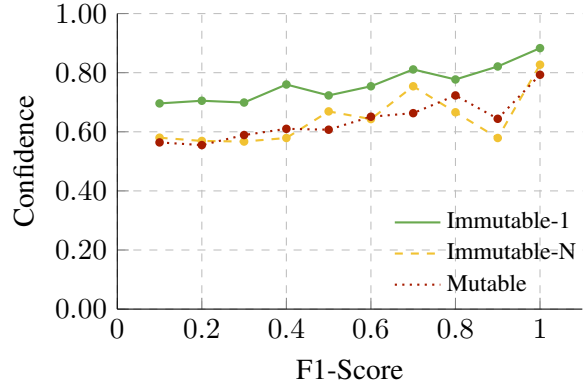

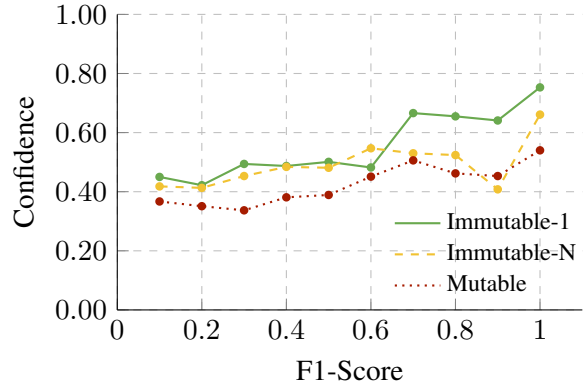
\begin{figure}[t]
\centering
\begin{tikzpicture}
\def\espacegraphique{45pt}
\begin{axis}[
    width=\linewidth, height=\height,
    name=int_cosine,
    grid=major, grid style={dashed,gray!50},
    xlabel=F1-Score, ylabel=Confidence,
    ymin=0.0, ymax=1.0, xmin=0.0, xmax=1.1,
    ytick={0.0,0.2,...,1.0},
    xtick={0.0,0.2,...,1.0},
    axis y line*=left, axis x line*=bottom,
    y tick label style={
        /pgf/number format/fixed,
        /pgf/number format/fixed zerofill,
        /pgf/number format/precision=2
    },
    legend pos=south east,
    legend cell align=left,
    legend columns=3,
    transpose legend,
    legend style={draw=none,outer sep=0pt,inner sep=0pt,
    fill=none, font=\footnotesize, yshift=0, xshift=5pt,
        /tikz/column 2/.style={column sep=5pt}
        },
    legend entries={
        Immutable-1,
        Immutable-N,
        Mutable
        }
    ]
    \addlegendimage{no markers, color_immutable1, solid, thick}
    \addlegendimage{no markers, color_immutablen, dashed, thick}
    \addlegendimage{no markers, color_mutable, dotted, thick}
    \addplot+[immutable1=color_immutable1] table[x=f1, y=immutable, col sep=comma]
    {data/f1_confidence_alpaca.csv};
    \addplot+[immutablen=color_immutablen] table[x=f1, y=immutable_n, col sep=comma]
    {data/f1_confidence_alpaca.csv};
    \addplot+[mutable=color_mutable] table[x=f1, y=mutable, col sep=comma]
    {data/f1_confidence_alpaca.csv};

\end{axis}

\end{tikzpicture}
\caption{Average confidence per F1-Score for Alpaca over all the queries in \mulam. 
}
\label{fig:f1_confidence_alpaca}
\vspace{-3mm}
\end{figure}

\begin{figure}[t]
\centering
\begin{tikzpicture}
\def\espacegraphique{45pt}
\begin{axis}[
    width=\linewidth, height=\height,
    name=int_cosine,
    grid=major, grid style={dashed,gray!50},
    xlabel=F1-Score, ylabel=Confidence,
    ymin=0.0, ymax=1.0, xmin=0.0, xmax=1.1,
    ytick={0.0,0.2,...,1.0},
    xtick={0.0,0.2,...,1.0},
    axis y line*=left, axis x line*=bottom,
    y tick label style={
        /pgf/number format/fixed,
        /pgf/number format/fixed zerofill,
        /pgf/number format/precision=2
    },
    legend pos=north east,
    legend cell align=left,
    legend columns=3,
    transpose legend,
    legend style={draw=none,outer sep=0pt,inner sep=0pt,
    fill=none, font=\footnotesize, yshift=0, xshift=5pt,
        /tikz/column 2/.style={column sep=5pt}
        },
    legend entries={
        Immutable-1,
        Immutable-N,
        Mutable
        }
    ]
    \addlegendimage{no markers, color_immutable1, solid, thick}
    \addlegendimage{no markers, color_immutablen, dashed, thick}
    \addlegendimage{no markers, color_mutable, dotted, thick}
    \addplot+[immutable1=color_immutable1] table[x=f1, y=immutable, col sep=comma]
    {data/f1_confidence_llama.csv};
    \addplot+[immutablen=color_immutablen] table[x=f1, y=immutable_n, col sep=comma]
    {data/f1_confidence_llama.csv};
    \addplot+[mutable=color_mutable] table[x=f1, y=mutable, col sep=comma]
    {data/f1_confidence_llama.csv};

\end{axis}

\end{tikzpicture}
\caption{Average confidence per F1-Score for Llama over all the queries in \mulam. 
}
\label{fig:f1_confidence_llama}
\vspace{-3mm}
\end{figure}
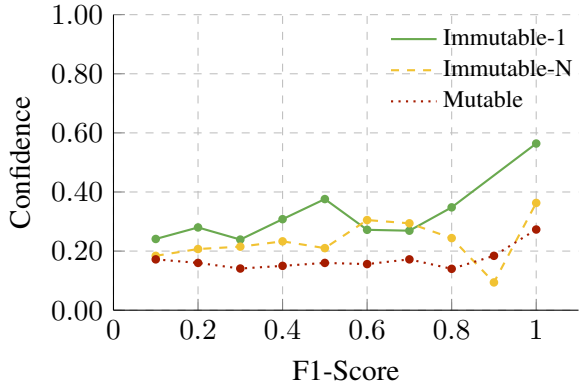

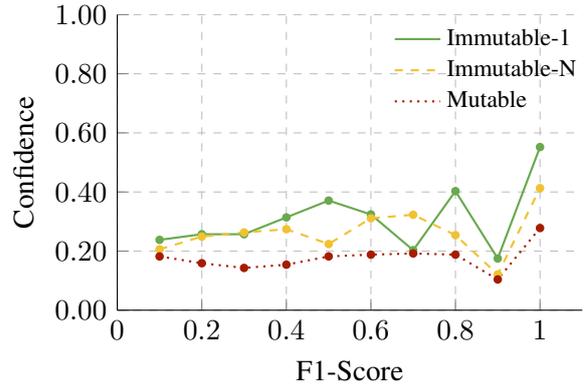
\begin{figure}[t]
\centering
\begin{tikzpicture}
\def\espacegraphique{45pt}
\begin{axis}[
    width=\linewidth, height=\height,
    name=int_cosine,
    grid=major, grid style={dashed,gray!50},
    xlabel=F1-Score, ylabel=Confidence,
    ymin=0.0, ymax=1.0, xmin=0.0, xmax=1.1,
    ytick={0.0,0.2,...,1.0},
    xtick={0.0,0.2,...,1.0},
    axis y line*=left, axis x line*=bottom,
    y tick label style={
        /pgf/number format/fixed,
        /pgf/number format/fixed zerofill,
        /pgf/number format/precision=2
    },
    legend pos=north east,
    legend cell align=left,
    legend columns=3,
    transpose legend,
    legend style={draw=none,outer sep=0pt,inner sep=0pt,
    fill=none, font=\footnotesize, yshift=0, xshift=5pt,
        /tikz/column 2/.style={column sep=5pt}
        },
    legend entries={
        Immutable-1,
        Immutable-N,
        Mutable
        }
    ]
    \addlegendimage{no markers, color_immutable1, solid, thick}
    \addlegendimage{no markers, color_immutablen, dashed, thick}
    \addlegendimage{no markers, color_mutable, dotted, thick}
    \addplot+[immutable1=color_immutable1] table[x=f1, y=immutable, col sep=comma]
    {data/f1_confidence_llama2.csv};
    \addplot+[immutablen=color_immutablen] table[x=f1, y=immutable_n, col sep=comma]
    {data/f1_confidence_llama2.csv};
    \addplot+[mutable=color_mutable] table[x=f1, y=mutable, col sep=comma]
    {data/f1_confidence_llama2.csv};

\end{axis}

\end{tikzpicture}
\caption{Average confidence per F1-Score for Llama2 over all the queries in \mulam. 
}
\label{fig:f1_confidence_llama2}
\vspace{-3mm}
\end{figure}

\begin{figure}[t]
\centering
\begin{tikzpicture}
\def\espacegraphique{45pt}
\begin{axis}[
    width=\linewidth, height=\height,
    name=int_cosine,
    grid=major, grid style={dashed,gray!50},
    xlabel=F1-Score, ylabel=Confidence,
    ymin=0.0, ymax=1.0, xmin=0.0, xmax=1.1,
    ytick={0.0,0.2,...,1.0},
    xtick={0.0,0.2,...,1.0},
    axis y line*=left, axis x line*=bottom,
    y tick label style={
        /pgf/number format/fixed,
        /pgf/number format/fixed zerofill,
        /pgf/number format/precision=2
    },
    legend pos=north east,
    legend cell align=left,
    legend columns=3,
    transpose legend,
    legend style={draw=none,outer sep=0pt,inner sep=0pt,
    fill=none, font=\footnotesize, yshift=0, xshift=5pt,
        /tikz/column 2/.style={column sep=5pt}
        },
    legend entries={
        Immutable-1,
        Immutable-N,
        Mutable
        }
    ]
    \addlegendimage{no markers, color_immutable1, solid, thick}
    \addlegendimage{no markers, color_immutablen, dashed, thick}
    \addlegendimage{no markers, color_mutable, dotted, thick}
    \addplot+[immutable1=color_immutable1] table[x=f1, y=immutable, col sep=comma]
    {data/f1_confidence_flant5.csv};
    \addplot+[immutablen=color_immutablen] table[x=f1, y=immutable_n, col sep=comma]
    {data/f1_confidence_flant5.csv};
    \addplot+[mutable=color_mutable] table[x=f1, y=mutable, col sep=comma]
    {data/f1_confidence_flant5.csv};

\end{axis}

\end{tikzpicture}
\caption{Average confidence per F1-Score for Falcon over all the queries in \mulam. 
}
\label{fig:f1_confidence_falcon}
\vspace{-3mm}
\end{figure}
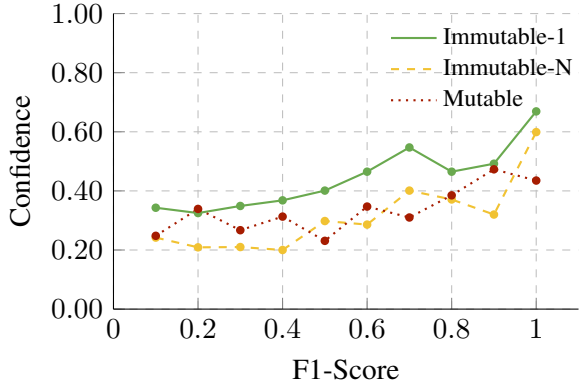

\section{Probe Classifier}\label{appendix:probe}
The relations that we use for training are:
\begin{itemize}
    \item Immutable-1: P103, P19, P159.
    \item Immutable-N: P27, P1412, P190.
    \item Mutable: P937, P286, P6.
\end{itemize}
The relations that we use for validation are:
\begin{itemize}
    \item Immutable-1: P20, P364.
    \item Immutable-N: P69, P101.
    \item Mutable: P108, P488.
\end{itemize}
The number of examples per split is (Immutable-1 / Immutable-N):
\begin{itemize}
    \item Train: 6230 / 6820.
    \item Validation: 5780 / 5910.
    \item Test: 35000 / 31410.
\end{itemize}

To obtain the LM representation and train the probe classifier we use one NVIDIA A100-40GB, each probe classifier takes a couple of hours to train.

\begin{table}[t]
\centering\scriptsize\setlength{\tabcolsep}{3pt}
\resizebox{0.48\textwidth}{!}{
\begin{tabular}{@{}ccccc@{}}
\toprule
\multicolumn{1}{l}{\textbf{}} & \multicolumn{2}{c}{\textbf{Imm-N}} & \multicolumn{2}{c}{\textbf{Imm-1}} \\ \cmidrule(l){2-5} 
\textbf{Percentile} & \textbf{Immutable-N} & \textbf{Mutable} & \textbf{Immutable-1} & \textbf{Mutable} \\ \midrule
0 & 8 & 1414 & 350 & 1414 \\
1 & 408 & 1262 & 840 & 1167 \\
2 & 609 & 1052 & 964 & 840 \\
3 & 439 & 1338 & 831 & 1262 \\
4 & 379 & 1330 & 859 & 1236 \\
5 & 519 & 1078 & 732 & 994 \\
6 & 804 & 930 & 856 & 1140 \\
7 & 918 & 729 & 1051 & 860 \\
8 & 995 & 639 & 1071 & 818 \\
9 & 1153 & 498 & 1341 & 539 \\ \bottomrule
\end{tabular}
}
\caption{Examples counts per frequency bin for each probe classifier.}\label{table:freq_counts_classifiers}
\end{table}

\subsection{Classifiers Performance vs Frequency}\label{appendix:freq_prob_plots}
Given the frequencies described in \ref{appendix:mulan_freq}, we conducted additional analysis to investigate whether the different behaviors we find are only due to differences in frequency. We analyze the classifiers accuracy on different frequency bins, in order to asses if it is simply learning to classify frequency features as opposed to mutability features.

\begin{figure*}[h!]
\centering
\begin{tikzpicture}
\def\espacegraphique{75pt}
\begin{axis}[
    width=0.95*\linewidth, height=6.5cm,
    name=int_cosine,
    grid=major, grid style={dashed,gray!50},
    xlabel=Bins, ylabel=Accuracy,
    ymin=0.55, ymax=1.0, xmin=0, xmax=9,
    ytick={0.55,0.65,...,1.0},
    xtick={0,1,...,9},
    axis y line*=left, axis x line*=bottom,
    y tick label style={
        /pgf/number format/fixed,
        /pgf/number format/fixed zerofill,
        /pgf/number format/precision=2
    },
    legend pos=north east,
    legend cell align=left,
    legend columns=3,
    transpose legend,
    legend style={draw=none,outer sep=0pt,inner sep=0pt,
    fill=none, font=\footnotesize, yshift=0, xshift=5pt,
        /tikz/column 2/.style={column sep=5pt}
        },
    legend entries={
        \alpaca{},
        \falcon{},
        \falconinst{},
        \llamaone{},
        \llamatwo{},
        \llamachat{}
        },
    cycle list name=exotic
    ]

    \addplot table[x=percentile, y=alpaca, col sep=comma]
    {data/acc_bins_classifier_1_n.csv};
    
    \addplot table[x=percentile, y=falcon, col sep=comma]
    {data/acc_bins_classifier_1_n.csv};
    
    \addplot table[x=percentile, y=falconinst, col sep=comma]
    {data/acc_bins_classifier_1_n.csv};

    \addplot table[x=percentile, y=llamaone, col sep=comma]
    {data/acc_bins_classifier_1_n.csv};

    \addplot table[x=percentile, y=llamatwo, col sep=comma]
    {data/acc_bins_classifier_1_n.csv};

    \addplot table[x=percentile, y=llamachat, col sep=comma]
    {data/acc_bins_classifier_1_n.csv};

\end{axis}

\end{tikzpicture}
\caption{Accuracy of the MDL classifier per percentile for all six models on the \textbf{Immutable-N} types of relations.}
\label{fig:acc_bins_classifier_1_n}
\vspace{-3mm}
\end{figure*}
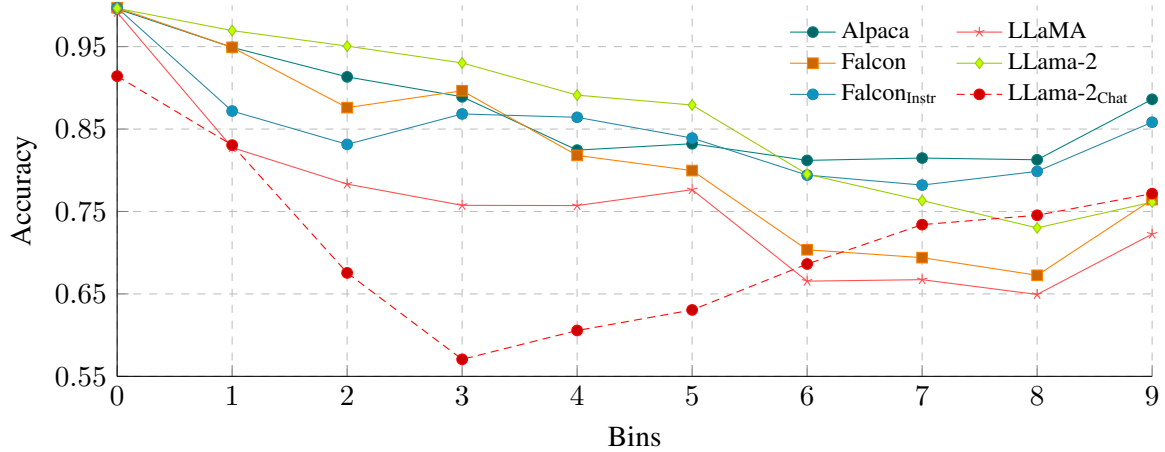

The distribution of classes per bin is presented in Table \ref{table:freq_counts_classifiers} and the accuracy for each bin and model is presented in Figure \ref{fig:acc_bins_classifier_1_1} and Figure \ref{fig:acc_bins_classifier_1_n} for the Imm-1 and Imm-N classifiers respectively. We observe from Table \ref{table:freq_counts_classifiers} that the classes (immutable/mutable) are roughly balanced in the different bins, but we do tend to have more mutable examples in the least frequent bins and more immutable in the higher-frequency bins. From the plots in Figure \ref{fig:acc_bins_classifier_1_1} and Figure \ref{fig:acc_bins_classifier_1_n} we see that the performance differs across bins, we obtain high accuracy (>90\%) in lower and upper bins, while mid bins accuracy is around 85\% for the Imm-1 classifier and around 80\% for the Imm-N classifier (it differs somewhat per model). However, we note that the performance of the classifiers cannot be explained only by frequency. For the Imm-N classifier e.g. in the 3rd bin, there are 609 immutable and 1052 mutable examples. Depending on the model, the classifier obtains between 70-95\% accuracy, whereas a random classifier using only the frequency would obtain ~60\%. The same goes with a higher bin, where there are 995 immutable and 639 mutable examples, and the accuracy is still ranging between 65-85\% whereas a classifier using only the frequency would obtain ~55\%.
In the same line for the Imm-1 classifier, we see that in the 3rd bin there are 964 immutable-1 and 840 mutable examples, and the models get around 85-95\% accuracy, way above a classifier using simple frequency (where low frequent examples would be considered mutable, therefore getting ~46\% accuracy). Similarly, in higher bins (9th), there are 1071 and 818 immutable-1 and mutable examples respectively, and the classifiers obtain 75-85\% accuracy, indicating that they are not simply considering all these high frequent examples mutable.

\begin{figure*}[h!]
\centering
\begin{tikzpicture}
\def\espacegraphique{75pt}
\begin{axis}[
    width=0.95*\linewidth, height=6.5cm,
    name=int_cosine,
    grid=major, grid style={dashed,gray!50},
    xlabel=Bins, ylabel=Accuracy,
    ymin=0.70, ymax=1.0, xmin=0, xmax=9,
    ytick={0.7,0.75,...,1.0},
    xtick={0,1,...,9},
    axis y line*=left, axis x line*=bottom,
    y tick label style={
        /pgf/number format/fixed,
        /pgf/number format/fixed zerofill,
        /pgf/number format/precision=2
    },
    legend pos=north east,
    legend cell align=left,
    legend columns=3,
    transpose legend,
    legend style={draw=none,outer sep=0pt,inner sep=0pt,
    fill=none, font=\footnotesize, yshift=0, xshift=5pt,
        /tikz/column 2/.style={column sep=5pt}
        },
    legend entries={
        \alpaca{},
        \falcon{},
        \falconinst{},
        \llamaone{},
        \llamatwo{},
        \llamachat{}
        },
    cycle list name=exotic
    ]

    \addplot table[x=percentile, y=alpaca, col sep=comma]
    {data/acc_bins_classifier_1_1.csv};
    
    \addplot table[x=percentile, y=falcon, col sep=comma]
    {data/acc_bins_classifier_1_1.csv};
    
    \addplot table[x=percentile, y=falconinst, col sep=comma]
    {data/acc_bins_classifier_1_1.csv};

    \addplot table[x=percentile, y=llamaone, col sep=comma]
    {data/acc_bins_classifier_1_1.csv};

    \addplot table[x=percentile, y=llamatwo, col sep=comma]
    {data/acc_bins_classifier_1_1.csv};

    \addplot table[x=percentile, y=llamachat, col sep=comma]
    {data/acc_bins_classifier_1_1.csv};

\end{axis}

\end{tikzpicture}
\caption{Accuracy of the MDL classifier per percentile for all six models on the \textbf{Immutable-1} types of relations.}
\label{fig:acc_bins_classifier_1_1}
\vspace{-3mm}
\end{figure*}
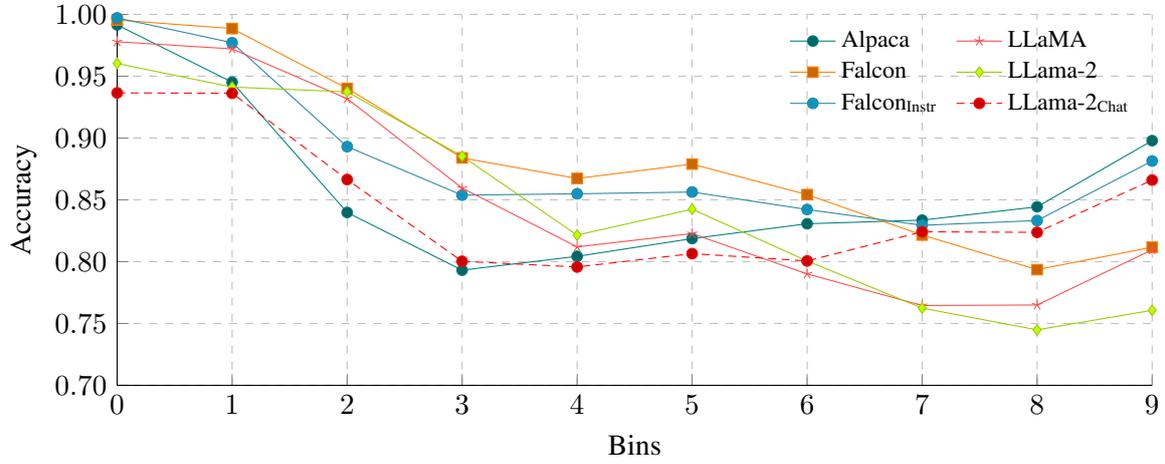

\section{MDL Experimental Setup}\label{appendix:mdl}
We train the probes using the Adam optimizer \citep{kingma2014adam}, with weight decay 0.01 and the default learning rate in the HuggingFace trainer (5e-5). We use a linear learning rate scheduler with warm-up ratio of \{0.0, 0.1, 0.2\}.\footnote{Defined specifically for each probe by using the one that gives the best validation accuracy when the probe is trained on all the data (on \(t_S\)).} We use early stopping with patience 4 and we train for a maximum of 12 epochs. 

\section{Updates Data}\label{appendix:updates}
After the selection process described in the Edits (Q3) section we obtain the following number of examples per model: 177 (\llamaone{}), 406 (\alpaca{}), 182 (\llamatwo{}), 2138 (\llamachat{}), 136 (\falcon{}), 583 (\falconinst{}).

\end{document}